# An All-Pair Quantum SVM Approach for Big Data Multiclass Classification

Arit Kumar Bishwas[a, *], Ashish Mani[b], Vasile Palade[c]

[a] Department of Information Technology, Noida, India, aritkumar.official@gmail.com
[b] Department of EEE, Amity University, Noida, India, amani@amity.edu
[c] Faculty of Engineering and Computing, Coventry University, Coventry, UK, vasile.palade@coventry.ac.uk

**Abstract**

In this paper, we have discussed a quantum approach for the all-pair multiclass classification problem. We have shown that the multiclass support vector machine for big data classification with a quantum all-pair approach can be implemented in logarithm run time complexity on a quantum computer. In an all-pair approach, there is one binary classification problem for each pair of classes, and so there are $k\,(k-1)/2$ classifiers for a $k-$class problem. As compared to the classical multiclass support vector machine that can be implemented with polynomial run time complexity, our approach exhibits exponential speed up in the quantum version. The quantum all-pair algorithm can be used with other classification algorithms, and a speed up gain can be achieved as compared to their classical counterparts.

*Keywords:* Multiclass Classification, SVM, Quantum Algorithm

**1. Introduction**

The recent technological advancements led us to depend on huge volume of data and it's mining to extract useful information. Many challenging problems nowadays require effective machine learning techniques. Performing any machine-learning task with a large volume of data bears high computational cost. Machine learning (ML) deals with the development of algorithms that can learn from and make predictions using data. At present, one of the most important challenges in machine learning is to deal with very large datasets (the so-called Big Data). ML tasks can be classified as supervised and unsupervised learning [1-4]. Clustering is an unsupervised learning process where the task is to find patterns with unlabeled data, whereas classification is a supervised learning process in which labeled (classified) data are used for training purpose, and then new data instances are classified.

Support vector machines (SVM) have been widely used as a binary classifier, but in recent years, multiclass support vector machines (MSVM) are also being widely studied, which classifies vectors into multiple sets with the help of trained oracles [5]. In binary SVM, the training activity is implemented by constructing an optimal hyperplane. This hyperplane divides the input data set into two groups, which is either in the original feature space or in a higher-dimensional kernel space. Many approaches have been proposed for constructing multiclass support vector machines with the help of a binary SVM, one of the most popular being the *all-pair* approach [6]. In this *approach,* there is one binary classification problem for each pair of classes, and therefore we construct $(k\,(k-1)/2\,)$ classifiers. Each classifier is a binary classifier and is trained with specific associated training examples. For the training purpose, we have to create $(k(k-1)/2)$ sets of training examples. Suppose, $M$ is the number of total training examples and $M^{(f,s)}$ is the training set for the classifier $h^{(f,s)}$. Where, $f = 1,2,…,k$, and $s = 1,2,…,k$. For example, for the training of the classifier $h^{(1,2)}$ we prepare a training set which contains only those examples from $M$ that have class values as $1\ or\ 2$. Now, each classifier predicts a class value for an unseen query state and store the value in a list $V$. Then by applying a voting mechanism on $V$, we choose the class value which appeared a maximum number of times in the list $V$.

In [7], *Rebentrost et. al.* demonstrated a quantum version of a support vector machine for binary classification. They have achieved logarithm time complexities for both the training and classification stages, so an exponential speedup gain (polynomial to logarithm time complexity) as compared to any classical counterpart. However, the approach does not support multiclass classification.

In [8], we have developed a quantum version of *one-against-all* technique to handle the quantum multiclass classification problem. Now further extending our work, in this paper we have proposed a quantum algorithm for multiclass classification by using a quantum version of an *all-pair* approach. We have demonstrated that the quantum multiclass approach can be implemented with logarithm time complexity as compared to polynomial time complexity in a classical multiclass approach. We have used the technique mentioned in [7] to construct the binary quantum SVM as a base, and then transform it for a multiclass quantum SVM by using the quantum *all-pair* technique. In the proposed approach, we first formulate $k(k-1)/2$ quantum binary least square support vector machines. Each classifier is trained with a specific data set. This specific data set contains training data sets corresponding to the two classes specifically. Suppose, $c_{ij}$ is a classifier then this classifier is trained with the training examples which are associated with the class values $i$ and $j$. The $k(k-1)/2$ classifiers operates as a quantum query operator to classify the state. Subsequently a quantum voting mechanism is applied for predicting the class of data-set. Once all the classifiers classify the state, we apply a quantum mechanically voting mechanism to predict the class value.

## 2. Quantum Binary Least Square Support Vector Machine

Given a training example set $\{(x_1, y_1), (x_2, y_2), \dots, (x_M, y_M)\}$, a support vector machine is a hyperplane that separates a set of a set of correctly-identified object instances from a set of incorrectly-identified object instances with maximum margin. Here, $x_i \in \mathbb{R}^d$ are input data objects, and $y \in \{1, -1\}$ are binary classes associated with the data objects. $M$ is the number of training instances, and $d$ is the number of features associated with a single data objects. The decision function in SVM is defined as

$$f(x) = sgn(w^T \emptyset(x) + b) \tag{1}$$

Where $x$ is the training instances, $w$ is the normal vector to the hyperplane, $f(x)$ is the predicted output, and $b$ is the offset of the hyperplane from the origin. Further, we solve the dual formulation of the optimization problem. We use Lagrangian multipliers to obtain the dual problem which accommodate constrains in the minimization problem. The dual problem becomes the following quadratic programming problem:

$$max: -\frac{1}{2}\sum_{i=1}^{M}\sum_{j=1}^{M}\alpha_i y_i \alpha_j y_j K(x_i, x_j) + \sum_{i=1}^{M}\alpha_i, \tag{2}$$

subject to $w = \sum_{i=1}^{M}\alpha_i y_i \emptyset(x_i), \sum_{i=1}^{M}\alpha_i y_i = 0, 0 \leq \alpha_i \leq c, i = 1, \dots, M$

Where, $K(x_i, x_j)$ is the kernel function and $\alpha_i \geq 0$ is the Lagrangian multiplier. The optimal point is in the saddle point of the Lagrangian function, which we obtain by taking the partial derivation of $w, b$, and $\xi$. $c$ is a regularization cost parameter.

A least square support vector machine translates an optimization problem into a set of linear equations. The least square version is obtained by reformulating the minimization problem as follows:

$$min: \frac{\mu}{2}w^T w + \frac{\zeta}{2}\sum_{i=1}^{M}e_{c,i}^2, \tag{3}$$

which subject to the following equality constraints:

$$y_i[w^T \emptyset(x_i) + b] = 1 - e_{c,i}, i = 1, \dots, M \tag{4}$$

Where $\mu$ and $\zeta$ are the hyperparameters to tune the amount of regularization versus the sum of squared error. Here, $e_{c,i}$ is the slack variable. We obtain the following least squares problem:

$$\begin{bmatrix} 0 & 1^T \\ 1 & K + \gamma^{-1}I \end{bmatrix}\begin{bmatrix} b \\ \alpha \end{bmatrix} = \begin{bmatrix} 0 \\ y \end{bmatrix}, \tag{5}$$

where $\gamma = \frac{\zeta}{\mu}, K \in \mathbb{R}^{M \times M}, K_{ij} = K(x_i, x_j)$.

The complexity of solving the least square formulation is $O(M^3)$, and the matrix calculation takes $O(M^2 d)$ times. Therefore, the overall complexity of the classical binary SVM is $O(M^2(M + d))$.

The system of linear equations involves kernel matrix for solving the problem. We can solve the system of linear equations much faster by speeding up the kernel matrix calculations and solving the least squares formulation. These speedups are possible in the quantum paradigm, hence the runtime of the quantum binary SVM becomes $O(logMd)$ [7]. The central ideas of the quantum binary SVM are as follows (referring the discussion in [9]):

a. Quantum data preparation by using quantum random access memory (QRAM)
b. Speedup gain in the kernel matrix calculation
c. Speedup gain in the least squares dual formulation

## 2.1. Quantum data preparation by using QRAM

The classical $d$ dimensional complex vector is defined in quantum paradigm with $log_2 d$ qubits onto quantum states in quantum random access memory (QRAM) [10-16], which takes only $O(log_2 d)$ steps to query the memory for reconstructing a state.

The QRAM allows accessing the data in the quantum parallel manner by performing memory access in coherent quantum superposition [17]. In a QRAM, the output registers and the address registers are composed of qubits. The addresses are in superposition in the address register. The QRAM returns a superposition of correlated information in a data register $dr$:

$$\sum_j \psi_j |j\rangle_{R_{Ad}} \rightarrow \sum_j \psi_j |j\rangle_{R_{Ad}} |D_j\rangle_{dr} \qquad (6)$$

Where $R_{Ad}$ is an address register and it contains a superposition of addresses $\sum_j \psi_j |j\rangle_{R_{Ad}}$. $D_j$ is the $j^{th}$ memory cell content. In QRAM, it takes only $O(log\, Md)$ operations to access the data and uses $O(Md)$ resources.

## 2.2. Speedup gain in the kernel matrix calculation

Kernel matrix calculation involves dot product evaluation. A quantum kernel matrix is prepared by using the inner product evaluation process [18] for least square reformulation. The inverse of the normalized kernel matrix is computed and then carried out the exponentiation of $\hat{K}^{-1}$ [19]. Here, $\hat{K}$ is the normalized kernel matrix.

To calculate a dot product of two training instances, first, we generate two quantum states $|\psi\rangle$ and $|\varphi\rangle$ with an ancilla variable. Then estimate the sum of the squared norms of the two training instances. At last, compare the two training instances and perform a projective measurement on the ancilla alone.

Considering a linear kernel $K_{linear}$, we calculate the dot product $x_i^T x_j = \left(\frac{|x_i|^2+|x_j|^2-|x_i-x_j|^2}{2}\right)$ in the kernel. We construct the quantum state $|\psi\rangle = \frac{1}{\sqrt{2}}(|0\rangle|x_i\rangle + |1\rangle|x_j\rangle)$ with the help of QRAM and also estimate the state $|\varphi\rangle = \frac{1}{(|x_i|^2+|x_j|^2)}(|x_i||0\rangle - |x_j||1\rangle)$. Now, we make the quantum state $\frac{1}{\sqrt{2}}(|0\rangle - |1\rangle) \otimes |0\rangle$ evolve with the Hamiltonian

$$H = (|x_i||0\rangle\langle 0| + |x_j||1\rangle\langle 1|) \otimes \sigma_x \qquad (7)$$

which results in the following state

$$\frac{1}{\sqrt{2}}(\cos(|x_i|t)|0\rangle - \cos(|x_j|t)|1\rangle) \otimes |0\rangle - \frac{i}{\sqrt{2}}(\sin(|x_i|t)|0\rangle - \sin(|x_j|t)|1\rangle) \otimes |1\rangle \qquad (8)$$

Measuring the ancilla bit with $t$ ($|x_i|t, |x_j|t \ll 1$), the complexity of constructing $|\varphi\rangle$ and $(|x_i|^2 + |x_j|^2)$ with accuracy $\epsilon$ is $O(\epsilon^{-1})$. We perform a swap test on the ancilla alone with $|\psi\rangle$ and $|\varphi\rangle$. In a swap test, if the two states are equal, then the measurement will be zero otherwise, it measures one.

Thus, the complexity of evaluating a single dot product $x_i^T x_j$ with QRAM is $O(\epsilon^{-1} log d)$.

*2.3. Speedup gain in the least squares dual formulation*

The improvement in computational complexity also benefits the training phase due to quantum implementation. The speedup gain is possible because of the quantum mechanically implementation of matrix inversion algorithm [20], exponentially faster eigenvector formulation in non-sparse density matrices [20], and simulating sparse matrices [21].

To solve $\hat{F}(|b, \vec{\alpha}\rangle) = |\vec{y}\rangle$, the matrix exponential of $\hat{F}$:

$$\hat{F} = \begin{bmatrix} 0 & 1^T \\ 1 & K + \gamma^{-1}I \end{bmatrix} \tag{9}$$

Where, $\hat{F} = \frac{F}{trF}$ is the normalized $F$, and $trF$ is a trace of $F$. We write $\hat{F}$ as $\hat{F} = \frac{(J+K+\gamma^{-1}I)}{trF}$. Where, $J = \begin{pmatrix} 0 & 1^T \\ 1 & 0 \end{pmatrix}$ is a star graph, and $K$ is the non-sparse kernel matrix. By using the Lie product formula, we obtain the following exponentiation

$$e^{-i\hat{F}\Delta t} = e^{-i\Delta tI/trF} e^{-iJ\Delta t/trF} e^{-iK\Delta t/trF} + O(\Delta t^2). \tag{10}$$

With multiple copies of density matrix $\rho$, it is possible to implement $e^{-i\rho t}$ [13]. We compute the matrix inverse $\hat{K}^{-1}$, where, $\hat{K}$ is a non-parse normalized Hermitian matrix. The run time complexity of the exponentiation is therefore $O(logd)$. We evaluate $e^{-i\hat{K}\Delta t}$ as

$$e^{-i\mathcal{L}_{\hat{K}}\Delta t}(\rho) \approx \rho - i\Delta t[\hat{K}, \rho] + O(\Delta t^2), \tag{11}$$

where, $\mathcal{L}_{\hat{K}} = [\hat{K}, \rho]$.

Using (11), we obtain the eigenvectors and eigenvalues by doing quantum phase estimation. Also, $|b, \vec{\alpha}\rangle$ is obtained by inverting the eigenvalues. So, the overall run time complexity for training is $O(\log Md)$.

**3. Earlier Work - Quantum Multiclass Classification with Quantum One-Against-All Approach**

In this section, we have briefly discussed our earlier work on multiclass classification. We have developed a quantum version of multiclass classification algorithm using quantum one-against-all approach (this algorithm has been developed previously by us) [8]. In this method, we first construct and train $k$ quantum binary classifiers [7]. Each of these quantum binary classifiers then classify a given quantum query state $|\vec{x}\rangle$ with some probability value. These values are stored quantum mechanically in the list $|V_{qn}\rangle$ into the QRAM for further processing. $|V_{qn}\rangle$ is the quantum superposition state of all the class probability values in QRAM. Then the quantum version of one-against-all algorithm runs on $|V_{qn}\rangle$ to find the class for which the corresponding classifier's probability confidence score is highest. The outcome produces the desired class by performing the quantum measurement operation (explained in the *Algorithm 1*). The quantum version of the one-against-all algorithm is given below:

---
**ALGORITHM 1:** Quantum One-Against-all Algorithm
---

1. *initialize index = any randon element, $1 \leq index \leq k$*
2. *initialize $|V_{qn}\rangle$ as the vector of all classified class probabilities in QRAM*
3. *while (total running time $< [O(\sqrt{k}) + O(log^2 k)])$*
4.     *initialize the memory as $|\Psi\rangle = \frac{1}{\sqrt{k}}\sum_r |r\rangle |index\rangle$*
5.     *GROVER-QUANTUM-SEARCH $(|\Psi\rangle, |V_{qn}\rangle, index)$*
6.     *if $(|V_{qn}\rangle[index_1] > |V_{qn}\rangle[index])$*
7.         *index = $index_1$*
8. *return index*

At first, we initialize the index, *index,* with a randomly and uniformly chosen value from the vector $|V_{qn}\rangle$. The *while* loop terminates when the total running time is greater than or equal to $(O(\sqrt{k}) + O(\log^2 k))$. We then initialize the memory, and call the function GROVER-QUANTUM-SEARCH ($|\Psi\rangle, |V_{qn}\rangle$, index) to apply the *Grover's Quantum Search*.

*Grover's Quantum Search* method uses Grover's Algorithm. Grover's Algorithm solves a general search problem in which there are $N$ elements that can be represented by $n$ basis states in the Hilbert Space i.e. $N \leq 2^n$, where $N$ and $n$ are both positive integers. Let $HS = \{0,1\}^n$ and let $O_f : HS \rightarrow \{0,1\}$, where '$O_f$' represents the oracle, which returns the answer when sampled but no other information is known about '$O_f$'. Using this framework along with quantum operators, the target state $|t_s\rangle$, which is a basis state in $HS$ such that $O_f(t_s) = 1$, is to be found. In the proposed method, '$O_f$' would return 1 for all cases where $|V_{qn}\rangle[r]$ is larger than $|V_{qn}\rangle[index]$, otherwise zero.

*Grover's Quantum Search* magnifies the amplitudes of all the indices whose corresponding vector values are greater than the threshold value $|V_{qn}\rangle[index]$ and marked them by satisfying the condition $|V_{qn}\rangle[r] > |V_{qn}\rangle[index]$. Once the amplitude of these elements has been magnified, we perform a measurement on $|r\rangle$ to obtain a new threshold index $index_1$. At the end, it returns the $index$ for the largest success probability value, which defines the classified class. Here, the iteration time $[O(\sqrt{k}) + O(\log^2 k)]$ in the algorithm is defined as $22.5\sqrt{k} + 1.4\log^2 k$ [22]. The probability for which the $|V_{qn}\rangle[index]$ holds the maximum value, which is our required index value in $|V_{qn}\rangle$ for predicting the associated class, is at least,

$$\left[1 - \frac{1}{2^{[O(\sqrt{k}) + O(\log^2 k)]}}\right], \tag{12}$$

where here $log$ is the binary logarithm.

Let us now discuss the upper bound, $22.5\sqrt{k} + 1.4\log^2 k$, of the algorithm. In the above *Algorithm 1*, the *step 4* takes $\log k$ time steps. The *steps 1, 2, 6, 7 & 8* are not counted for consideration. Let us assume that the algorithm runs enough to find the maximum value in $|V_{qn}\rangle$. The algorithm looks for the maximum among $m$ items which are greater than $|V_{qn}\rangle[index]$. The following *Lemma 1* [22] states that the probability with which an element in $|V_{qn}\rangle[index]$ will be chosen as the threshold is the inverse of the rank $r$ of the element. This probability is also independent of the size of the $|V_{qn}\rangle$ $index$].

**Lemma 1** *When an infinite algorithm searches among m elements, the probability $p(m, r)$ of choosing the index of the element of rank r will be*

$$p(m,r) = \begin{cases} \frac{1}{r} & ; if\ r \leq m \\ 0 & ; otherwise \end{cases}$$

As discussed in [22], the exponential search algorithm to find the index of a marked item (in this case, $m$ items are marked) among $k$ items is at most $4.5\sqrt{k/m}$. We now deduce the expected time for finding the maximum probability value in $|V_{qn}\rangle[index]$. Consider the following *Lemma 2* [22] which helps us to find the upper bound of the algorithm's iteration time.

**Lemma 2** *The estimated total time required by the infinite algorithm is at most $z = 11.25\sqrt{k} + 0.7\log^2 k$, before the index holds the index of the maximum.*

In the *Algorithm 1*, $|V_{qn}\rangle[index]$ holds the maximum value with probability at least ½ after at most $2z$ iterations. Therefore, the overall upper bound of the iteration time is $2z = 2(11.25\sqrt{k} + 0.7\log^2 k) = 22.5\sqrt{k} + 1.4\log^2 k$.

## 4. Quantum Multiclass SVM Classification with Quantum All-Pair Approach

In this section, we have discussed the proposed quantum multiclass SVM classification algorithm. We first prepared the dataset for $k(k-1)/2$ classifiers, where $k$ is the number of classes. Each classifier in the set of $k(k-1)/2$ classifiers is a quantum binary SVM classifier. We then, train $k(k-1)/2$ quantum binary classifiers with the associated dataset. During the prediction phase, each quantum binary classifier predicts a class. We then apply a quantum version of all-pair approach (which is developed by us to achieve the goal of multiclass classification in this research work). All-pair approach helps in identifying the class which occurred a maximum number of times in the $k(k-1)/2$ classification set. The identified class is the final predicted class for the multiclass classification problem. In the next section, we have discussed the computational complexity of the proposed algorithm.

As per our earlier discussion, the quantum binary support vector machine parameters are determined as follows:

$$\hat{F}(|b, \vec{\alpha}\rangle) = |\vec{y}\rangle \tag{13}$$

$$\Rightarrow |b, \vec{\alpha}\rangle = \hat{F}^{-1}(|\vec{y}\rangle) \tag{14}$$

In the similar context, we formulate the multiclass case as follows:

$$\hat{F}_j(|b_j, \vec{\alpha}_j\rangle) = |\vec{y}_j\rangle; j = 1, 2, 3, \ldots k(k-1)/2. \tag{15}$$

$$\Rightarrow |b_j, \vec{\alpha}_j\rangle = \vec{F}_j^{-1}(|\vec{y}_j\rangle) \tag{16}$$

Where, $\vec{y}_j = (y_{j1}, \ldots, y_{jM})^T$, $b_j$ is the biasing, $\vec{\alpha}_j = (\alpha_{j1}, \ldots, \alpha_{jM})^T$ is non parse & act as the distance from the optimal margin for the $jth$ classifier, and $k$ is the number of classes.

To solve the above formulations (equations 15 & 16) we have used quantum matrix inversion algorithm [20]. The matrix $\hat{F}_j$ can be decomposed as

$$\hat{F}_j = \frac{(J_j + K_j + \gamma_j^{-1}\mathbb{I})}{trF_j} \tag{17}$$

with $K_j$ as symmetric kernel matrices, $\gamma_j$ determines the relative weight of training error & least square SVM objectives, and $J_j = \begin{pmatrix} 0 & \vec{1}^T \\ \vec{1} & 0 \end{pmatrix}$ [23]. Further, for performing the quantum matrix inversion such as $\vec{F}_j^{-1}$ we need to indorse $e^{\frac{-i\hat{F}_j\Delta t}{trF}}$ efficiently, and as the matrix $\hat{F}_j$ is not sparse so exponentiation can be achieved by using the techniques discussed in [24], therefore

$$e^{\frac{-i\hat{F}_j\Delta t}{trF}} = e^{\frac{-iJ_j\Delta t}{trF}} e^{\frac{-iK_j\Delta t}{trF}} e^{\frac{-i\mathbb{I}_j\Delta t}{trF}} + O(\Delta t^2), \tag{18}$$

the matrices $\gamma_1^{-1}\mathbb{I}, \gamma_2^{-1}\mathbb{I}, \gamma_3^{-1}\mathbb{I}, \ldots \gamma_{k(k-1)/2}^{-1}\mathbb{I}$ are technically trivial. $O(\Delta t^2)$ is the error.

Also, $|\tilde{y}_j\rangle = \sum_{l=1}^{M_j^{(f,s)}+1} \langle u_l^j|\tilde{y}|u_l^j\rangle$; $j = 1, 2, 3, \ldots, k(k-1)/2$; $f = 1, 2, \ldots, k$; $s = 1, 2, \ldots, k$; $f, s \subseteq k$ & $f \neq s$; $|u_l^j\rangle$ are eigenstates of $\hat{F}_j$ with corresponding eigenvalues $|\lambda_l^j\rangle$. The eigenvalues states are stored by using phase estimation.

$$|\tilde{y}_j\rangle |0\rangle \rightarrow \sum_{l=1}^{M_j^{(f,s)}+1} \frac{\langle u_l^j|\tilde{y}_j\rangle}{\lambda_l^j} u_l^j\rangle \tag{19}$$

$M_j^{(f,s)} \subset M$ is the training data set used for training the $j^{th}$ classifier, and $M_j^{(f,s)}$ contains only those training data set which is having $f$ and $s$ as classes.

$$|b_j, \vec{\alpha}_j\rangle = \frac{1}{\sqrt{C_j}}\left(b_j|0\rangle + \sum_{l=1}^{M_j^{(f,s)}} \alpha_l^j |l\rangle\right); j = 1, 2, \ldots, k(k-1)/2 \tag{20}$$

where, $C_j = b_j^2 + \sum_{l=1}^{M_j^{(f,s)}} (\alpha_l^j)^2$ and $C_1, C_2, C_3, \ldots C_{k(k-1)/2}$ are the SVM parameters for $k(k-1)/2$ different classifiers. $\alpha_l^j$ is the value of $\alpha_l$ for $j^{th}$ case. With this approach, we have constructed $k(k-1)/2$ quantum binary SVM classifiers. Each classifier uses the sample $M_j^{(f,s)} \subset M$ for training the $jth$ classifier, where $M_j^{(f,s)}$ is the training data set containing those examples which are associated with the class $f$ and $s$ only.

Once all the $k(k-1)/2$ quantum binary SVMs have been trained, the next step is to discuss the classification part. In this case, there are $k(k-1)/2$ classifiers, which will classify a quantum query state. Let $h_j^{(f,s)}$ be the classifier, where $f$ & $s$ represent positive and negative valued class data set respectively. Also, $h_j^{(f,s)} = -h_j^{(f,s)}$.

First of all, we construct $k(k-1)/2$ quantum states with training-data oracles:

$$|\tilde{u}_j\rangle = \frac{1}{\sqrt{(N_{\tilde{u}})_j}} \left( b_j |0\rangle |0\rangle + \sum_{l=1}^{M_j^{(f,s)}} \alpha_l |\vec{x}_l^j\rangle |l\rangle |\vec{x}_l^j\rangle \right) \qquad (21)$$

where,

$$(N_{\tilde{u}})_j = b_j^2 + \sum_{l=1}^{M_j^{(f,s)}} (\alpha_l^j)^2 (|\vec{x}_l^j|)^2 \, ; j = 1,2,3,\ldots,(k(k-1)/2), \qquad (22)$$

We also construct the $j$ quantum query states, considering a specific dataset $M_j^{(f,s)}$ for $(k(k-1)/2)$ classifiers:

$$|\tilde{x}_j\rangle = |\tilde{x}_{(f,s)}\rangle = \frac{1}{\sqrt{N_{\tilde{x}_j}}} \left( |0\rangle |0\rangle + \sum_{l=1}^{M_j^{(f,s)}} |\vec{x}| |l\rangle |\vec{x}\rangle \right) \qquad (23)$$

Where, $N_{\tilde{x}_j} = M_j^{(f,s)} |\vec{x}|^2 + 1$; $j = 1,2,3,\ldots,(k(k-1)/2)$; $f = 1,2,3,\ldots,k$; and $s = 1,2,3,\ldots,k$. Here, $|\tilde{x}_j\rangle$ is the constructed quantum query state which is associated with the $j th$ classifier. The $j^{th}$ classifier has been trained with the labeled data set which are associated with only $f$ and $s$ classes.

We use an ancilla to construct $k(k-1)/2$ quantum states $|\psi_j\rangle = \frac{1}{\sqrt{2}} (|0\rangle |\tilde{u}_j\rangle + |1\rangle |\tilde{x}_j\rangle)$; $j = 1,2,3,\ldots,(k(k-1)/2)$, and measure this ancilla in the state $|\phi\rangle = \frac{1}{\sqrt{2}} (|0\rangle - |1\rangle)$ (can be constructed by applying a Hadamad gate to $|1\rangle$). The $(k(k-1)/2)$ success probabilities of the measurement are

$$P_j^{(f,s)} = |\langle \psi_j | \phi \rangle|^2 = \frac{1}{2} (1 - \langle \tilde{u}_j | \tilde{x}_j \rangle); f \neq s, \qquad (24)$$

where, $j = 1,2,3,\ldots,(k(k-1)/2), f = 1,2,3,\ldots,k$, and $s = 1,2,3,\ldots,k$.

Here, the $P_j^{(f,s)}$ represents the probability of the measurement by $j^{th}$ classifier. The algorithm predicts either $f$ or $s$ class based on the conditions mentioned in the *Table 1*. Hence, $|\vec{x}\rangle$ will be classified as $+1$ or $-1$ based on the following conditions,

**Table 1. Probability Conditions for Classification**

| Conditions and ($f \neq s$) | $|\vec{x}\rangle$ is classified as | Classified Class |
|---|---|---|
| $P_j^{(f,s)} < \frac{1}{2}$ | $+1$ | $f$ |
| $P_j^{(f,s)} \geq \frac{1}{2}$ | $-1$ | $s$ |

Also, $\langle \tilde{u}_j | \tilde{x}_j \rangle$ is $O(1)$ costly when $\alpha_l$ is not sparse, and $P_j^{(f,s)}$ can be obtained by iterating $O\left(\frac{P_j^{(f,s)}\left(1-P_j^{(f,s)}\right)}{\epsilon^2}\right)$ associated with the $j^{th}$ classification with accuracy $\epsilon$.

Suppose that $V$ ($|V| = k(k-1)/2$) is a list which stores all $k(k-1)/2$ predicted classes. Each classifier $h_j^{(f,s)}$, $f \neq s$, updates the list $V$ with the classified classes as follows:

**ALGORITHM 2:** Storing Classified Values

```
1: initialize j = 1, f = 1, s = 1, and V[j]
2: while j ≤ (k(k-1)/2)
3:     if ((f ≠ s) & (P^(f,s) < 1/2))
4:         V[j] = f
5:     else if ((f ≠ s) & (P^(f,s) > 1/2))
6:         V[j] = s
7:     j = j + 1
8:     f = f + 1
9:     s = s + 1
10: return V
```

Once all the predicted classes are stored in $V$, we transform the classical data of list $V$ into the quantum form using quantum random access memory [10-16]. Let $|V_q\rangle$ be the quantum list which holds the transformed quantum form of data in quantum random access memory (QRAM). Therefore, all the values in $|V_q\rangle$ are in quantum superposition into the QRAM.

$$|V_q\rangle = \sum_{i=0}^{\{\frac{k(k-1)}{2}\}-1} \tau_i |V_{q_i}\rangle \tag{25}$$

Where $|V_q\rangle$ is the quantum superposition state of all the $\left\{\frac{k(k-1)}{2}\right\}$ predicted classes in QRAM, $\tau_i$ is the complex coefficient associated with the quantum state $|V_{q_i}\rangle$, and $|V_{q_i}\rangle$ represents the $i^{th}$ predicted class value in quantum form. There are $k(k-1)/2$ predicted classes in total, and the dimension of each of the objects in $V$ is 1 (only class value), therefore the total cost for this transformation is

$$O\left(\log\left(\frac{k(k-1)}{2}\right)\right) \sim O(\log k^2). \tag{26}$$

We now define a quantum *all-pair* function $f_{all-pair}(|V_q\rangle)$ which returns the desired predicted class. The following algorithm describes the quantum mechanically implemented function $f_{all-pair}(|V_q\rangle)$ for the *all-pair* approach:

---

**ALGORITHM 3:** Quantum All-Pair Algorithm

---

```
1: initialize class = any random element, 1 ≤ class_index ≤ k(k-1)/2
2: initialize |V_q⟩ as the vector of all classified classes
3: initialize frequency estimate s_{class_index} with any very small value
4: INITIAL-FREQUENCY-COUNT (class_index, s_{class_index})
5: while (total running time < O(log k))
6:     initialize the memory |C_i⟩ = Σ_j β_{i,j} |j⟩
7:     initialize the memory as
           |ψ_c⟩ = Σ_{i=0}^{k(k-3)/2} |i⟩ |C_i⟩|class_index⟩|s_{class_index}⟩
8:     QUANTUM-SEARCH (|ψ_c⟩, |V_q⟩, class_index, s_{class_index})
9:     MEASURE-REGISTER (|ψ_c⟩)
10:    if (s_{class_index_new} > s_{class_index} + ε/(k(k-1)))
11:        class_index = class_index_{new}
12:        s_{class_index} = s_{class_index_new}
13: return |V_q⟩[class_index]
```

In the above algorithm, at first, we initialize a random index variable *class_index*. The INITIAL-FREQUENCY-COUNT (*class_index*, $s_{class\_index}$) function calculates the initial frequency estimate $s_{class\_index}$ of *class_index*, then followed by initialized a count state $|C_i\rangle$. $|C_i\rangle$ defines the number of times the $i^{th}$ class occurs in vector $|V_q\rangle$ as a fraction of $\left(\frac{k(k-1)}{2}\right)$ classes. The state $|C_i\rangle$ is in a superposition of all the classes of $|V_q\rangle$, where each approximate count is a distribution over possible values of $C_i$ (in the classical sense). Here, the cardinality of $s_{class\_index}$ is $(k-1)$.

Next, we initialize a quantum state $|\psi_c\rangle$. The first register in $|\psi_c\rangle$ outputs the new class value $class\_index_{new}$. The second register outputs the new frequency estimate of $class\_index_{new}$. The function QUANTUM-SEARCH ($|\psi_c\rangle, |V_q\rangle, class\_index, s_{class\_index}$) performs a Grover's quantum search on $|V_q\rangle$ to find out the desired class. So technically, quantum search actually amplifies the amplitude of all the indices whose corresponding vector values are greater than the threshold value $|V_q\rangle [class\_index]$. Once the amplitude of these elements has been amplified, we perform a measurement to obtain a new threshold index $class\_index_{new}$.

The function MEASURE-REGISTER ($|\psi_c\rangle$) does a measurement of each register of the state $|\psi_c\rangle$. Based on the condition ($s_{class\_index\_new} > s_{class\_index} + \frac{\varepsilon}{k(k-1)}$), the $class\_index$ and $s_{class\_index}$ are updated. Here, $\varepsilon$ is the error during measurement. At the end, the algorithm returns the value $|V_q\rangle[class\_index]$ which is the desired predicted class. So the algorithm finds the class whose sampled frequency is an *ε-approximation* to that of the frequency value.

## 5. Computational Complexity Analysis

In a classical binary least square SVM, one of the primary computational cost is due to the calculation of the Lagrange multipliers $\alpha$ [25-26]. When extending the discussion to multiclass classification with least square SVM, the class set cardinality is considered in addition. The computational complexity for SVM is calculated by considering the total training and total classification costs. As we discussed in the earlier section 2, the overall computational cost for the classical binary SVM is $\approx O(M^3)$, which is a polynomial time complexity. When we consider the case of multiclass SVM ($k > 2$), the run time still appears to be polynomial (even including the cost of the classical *all-pair* algorithm, where there are $(k(k-1)/2)$ classifiers in total).

We have seen in [7], that the quantum binary class support vector machine exhibits exponential speed up as compared to the classical version of binary class support vector machine. We begin the runtime analysis of quantum binary support vector machine, later we will extend the discussion for the case of proposed quantum multiclass classification (with a quantum *all–pair* approach).

The kernel matrix preparation causes $O(\log(M_{max} d))$ run time costs, where $M_{max} = max\left\{|M_1^{(f,s)}|, |M_2^{(f,s)}|, ..., |M_{k(k-1)/2}^{(f,s)}|\right\} < |M|$; $f = 1, 2, 3, ..., k$; $s = 1, 2, 3, ..., k$; and $f \neq s$. In case of a single classifier, the number of time steps in phase estimation $T$ requires $O(t_0^2 \epsilon^{-1})$. Where $t_0^2$ is the total evolution time which determines the phase estimation error, and $\epsilon$ is the maximally error. Combining, we get $O(t_0^2 \epsilon^{-1} (\log(M_{max} d)))$ run time. Let's define a constant $\epsilon_{Kr}$ such that $\epsilon_{Kr} \leq |\lambda_l| \leq 1$, and an effective condition number $\kappa_{eff} = \epsilon_{Kr}^{-1}$. $\kappa_{eff}$ is used to employ the filtering procedure in phase estimation, referring [20]. By considering all the analysis, and iterating the algorithm for $O(\kappa_{eff})$ times for achieving a constant success probability of the post selection step, the overall run time is $O(\kappa_{eff}^3 \epsilon^{-3} (\log(M_{max} d)))$. So, we can scale it to $O(\log(M_{max} d))$. We have analyzed the computational complexity of a single quantum binary classifier. Extending the analysis for our multiclass case when there are $k(k-1)/2$ classifiers to be trained, and then to classify a quantum query state by each of these quantum binary classifiers.

We now analyze the run time complexity of the proposed quantum multiclass SVM with the quantum *all-pair* approach. The overall run time of the quantum multiclass SVM with quantum *all-pair* approach = *(total run time during training of all classifiers) + (total run time during classification) + (total run time of quantum all-pair approach)*.

As discussed, for a single quantum binary classifier the total run time (during training and classification phases) is $O(\log(M_{max} d))$. Hence, in case of $k(k-1)/2$ classifiers the total run time will be

$$O\left(\frac{k(k-1)}{2} (\log(M_{max} d))\right) \sim O(k^2 (\log(M_{max} d))). \tag{27}$$

In a quantum mechanically implementation of the *all-pair* approach, we are actually finding the mode amongst all the classifier's success probabilities. It is a kind of voting mechanism. The mentioned approach has been inspired by [27], where a quantum $\{\varepsilon, \delta\}$-FPRAS (Fully Polynomial-time Randomized Approximation scheme) algorithm has been discussed for mode finding. Also, the transformation of the classical data from list $V$ into quantum form in the list $V_q$ adds $O(logk)$ costs.

By using the quantum mechanical voting mechanism, we determine the modal value in $V_q$, which is actually the predicted class that occurred a most number of times in $V_q$. In this case, the upper bound of the modal value is bounded with $(k-1)$. The analysis shows that the run time of quantum mechanically implemented the *all-pair* algorithm is $O(k^3 \log(\delta^{-1})\varepsilon^{-1})$, which is a quadratic cost reduction as compared to the classical version of *all-pair*. Here, the approximate mode having a sampled frequency which is within the ratio $(1+\varepsilon)$ of the mode's frequency and at least with $(1-\delta)$ probability. The overall run time is therefore:

$$[O\ (k^2\ (\log(M_{max}\ d))) + O(k^3 \log(\delta^{-1})\ \varepsilon^{-1}) + O\ (logk)] \tag{28}$$

Let's now analyze the following cases:

*Case 1: When $k = 2$*

This is the case of quantum binary classification and in this case, the runtime will be

$$O\ (2^2\ (\log(M_{max}\ d))) + O(2^3 log(\delta^{-1})\varepsilon^{-1}) + O\ (log2) \cong O\ (\log(M_{max}\ d)) \tag{29}$$

*Case 2: When $2 < k \ll M_{max}$*

In this case, although $k > 2$ but still small as compare to $M_{max}$. Although, there is a significant impact to the run time by the factor value of $k$ (as compared to Case 1). So the run time is:

$$O\ (k^2\ (\log(M_{max}\ d))) + O(k^3 log(\delta^{-1})\varepsilon^{-1}) + O\ (logk) \tag{30}$$

*Case 3: When $k \to M_{max}$*

In this case, the impact of $k$ is very significant. It is always recommended to take a very large training set compared to the number of classes and always try to maintain $k \ll M_{max}$. The run time is

$$\lim_{k \to M_{max}} O\ (k^2\ (\log(M_{max}\ d))) + O(k^3 log(\delta^{-1})\varepsilon^{-1}) + O\ (logk) \tag{31}$$

$$\cong O\ (M_{max}^2 (\log(M_{max}\ d))) + O(M_{max}^3 log(\delta^{-1})\varepsilon^{-1} + O\ (logM_{max})) \tag{32}$$

We saw the clear quantum advantages with the proposed quantum multiclass classification and achieved the exponential speed up as compared to the classical version of the multiclass support vector machine.

## 6. Conclusion

In this paper, we have shown that a multiclass support vector machine can be implemented quantum mechanically with logarithm run time complexity whereas the known classical multiclass support vector machine has polynomial run-time complexity. We have analyzed and addressed the quantum multiclass SVM method with the quantum mechanically implemented *all-pair* approach. With the quantum *all-pair* approach, we have constructed k (k − 1)/2 classifiers quantum mechanically and allowed the classifiers to classify the given unseen quantum query state. Further, a voting scheme is applied where the class that appears a maximum number of times gets predicted. Moreover, we have determined the run time complexity of the proposed algorithm which shows significant speed up gain as compared to the classical counterpart.